\documentclass[journal]{IEEEtran}
 
\usepackage{hyperref}
\usepackage{amsmath,amsfonts,amssymb}
\usepackage{verbatim}
\usepackage[linesnumbered,ruled,vlined]{algorithm2e}%[ruled,vlined]{  
\usepackage{algpseudocode} 
\usepackage{graphicx}
\usepackage{textcomp}
\usepackage{multirow} 
\usepackage{epstopdf}
\usepackage{color}
\usepackage{threeparttable}
\usepackage[table,xcdraw]{xcolor}
\usepackage{subfigure}
\usepackage{changepage}
\usepackage{bm}
\usepackage{diagbox}
\usepackage{booktabs}
\usepackage[switch]{lineno}
\usepackage{pifont}
\usepackage{enumitem}
\newcommand{\mTa}{\mathbf{\Theta}}
\newcommand{\wt}{\mathbf{w}}
\newcommand{\bbC}{\mathbb{C}}
\newcommand{\bht}{\mathbf{h}}
\newcommand{\fd}{\mathrm{d}}
\newcommand{\fr}{\mathrm{r}}
\newcommand{\Gt}{\mathbf{G}}
\newcommand{\ie}{\textit{i}.\textit{e}.}
\newcommand{\eg}{\textit{e}.\textit{g}.}

\begin{document}
\title{Automated Metaheuristic Algorithm Design with Autoregressive Learning}

\author{Qi~Zhao,~
        Tengfei~Liu,~
        Bai~Yan,~
        Qiqi~Duan,~
        Jian~Yang,~
	and~Yuhui~Shi,~\IEEEmembership{Fellow,~IEEE}	

\thanks{This work is supported by Guangdong Basic and Applied Basic Research Foundation under Grant No. 2021A1515110024, Shenzhen Fundamental Research Program under Grant No. JCYJ20200109141235597, National Natural Science Foundation of China under Grants No. 61761136008, Shenzhen Peacock Plan under Grant No. KQTD2016112514355531, and Program for Guangdong Introducing Innovative and Entrepreneurial Teams under Grant No. 2017ZT07X386. \textit{(Corresponding author: Yuhui Shi)}}
\thanks{Q. Zhao, T. Liu, J. Yang, and Y. Shi are with the Department of Computer Science and Engineering, Southern University of Science and Technology, Shenzhen 518055, China (email: zhaoq@sustech.edu.cn; 12332470@mail.sustech.edu.cn; yangj33@sustech.edu.cn; shiyh@sustech.edu.cn).}
\thanks{B. Yan is with the Research Institute of Trustworthy Autonomous Systems, Southern University of Science and Technology, Shenzhen 518055, China (email: yanb@sustech.edu.cn).}
\thanks{Q. Duan is with the School of Computer Science and Technology, Harbin Institute of Technology, Shenzhen 518055, China (email:11749325@mail.sustech.edu.cn).}}

\maketitle
\begin{abstract}
Automated design of metaheuristic algorithms offers an attractive avenue to reduce human effort and gain enhanced performance beyond human intuition. Current automated methods design algorithms within a \textit{fixed} structure and operate from \textit{scratch}. This poses a clear gap towards fully discovering potentials over the metaheuristic family and fertilizing from prior design experience. To bridge the gap, this paper proposes an autoregressive learning-based designer for automated design of metaheuristic algorithms. Our designer formulates metaheuristic algorithm design as a sequence generation task, and harnesses an autoregressive generative network to handle the task. This offers two advances. First, through autoregressive inference, the designer generates algorithms with diverse lengths and structures, enabling to fully discover potentials over the metaheuristic family. Second, prior design knowledge learned and accumulated in neurons of the designer can be retrieved for designing algorithms for future problems, paving the way to continual design of algorithms for open-ended problem-solving. Extensive experiments on numeral benchmarks and real-world problems reveal that the proposed designer generates algorithms that outperform all human-created baselines on 24 out of 25 test problems. The generated algorithms display various structures and behaviors, reasonably fitting for different problem-solving contexts. Code will be released after paper publication.
\end{abstract}

% The network is trained to maximize the expected performance of the generated algorithm on a target problem(s).


\section{Introduction}
Metaheuristic is a stochastic search methodology that integrates gradient-free local improvement with high-level strategies of escaping from local optima \cite{glover2006handbook}. It involves a rich family of algorithms ranging from neighborhood search-based to evolutionary and swarm ones. These algorithms can be used for any problem with an available solution representation, solution quality evaluation, and notion of locality\footnote{Locality denotes the ability to generate neighboring solutions via a heuristically-informed function of one or more incumbent solutions \cite{swan2022metaheuristics}.} \cite{swan2022metaheuristics}. This has made metaheuristic algorithms dominant in solving hard problems that do not meet analytical algorithms' rigorous mathematical assumptions, e.g., convexity and differentiability.  

Given a target problem, carefully designing a metaheuristic algorithm over the rich choices of search components, algorithmic structures, and hyperparameters is necessary, in order to obtain good enough solutions. The design is usually human-made, criticized as laborious, untraceable regarding what motivate certain design decisions, and susceptible to accidental human bias, let alone that human designers are not always available \cite{zhao2023survey}. This motivates the automated design of metaheuristic algorithms \cite{zhao2023survey,stutzle2019automated,2020The}. 

The automated design could find either instantiations/variants of existing algorithms or unseen algorithms to fit for target problem-solving. This is different with automated algorithm selection \cite{2018Automated} and automated hyperparameter configuration \cite{schede2022survey}. The former selects off-the-peg algorithms from a portfolio; the latter configures hyperparameters of a given algorithm. The automated design can be conducted either \textit{offline} by a target distribution of problem instances or \textit{online} by the search trajectory \cite{swan2022metaheuristics}. This paper focuses on \textit{offline automated design}, which suits scenarios where one can afford a priori computational resource (for algorithm design) to subsequently solve many problem instances drawn from the target domain. 

Several methods have been introduced to offline automated design of metaheuristic algorithms. Representatives include SMAC \cite{hutter2011sequential,lindauer2022smac3}, irace \cite{lopez2016irace}, ParamILS \cite{hutter2009paramils}, and offline hyperheuristics \cite{burke2013hyper,pillay2018hyper}. SMAC uses Bayesian optimization to learn a surrogate function that models the mapping from algorithms to their performance on a target problem. irace proposes iterative racing to select desired algorithms for a target problem, subsequently using the desired algorithms to approximate the sampling distribution of algorithms. ParamILS employs iterative local search to exploit desired algorithms; the intensification and capping strategies are developed to save algorithm evaluations. Offline hyperheuristics adopt high-level heuristics/metaheuristics, \eg, genetic programming \cite{koza1994genetic}, to generate low-level metaheuristic algorithms. 

The above methods show two common points. First, they manipulate a fixed-length vectorized algorithm representation, \ie, a concatenation of categorical indexes of algorithmic components and their conditional hyperparameters; a specific algorithm template is required to determine the components' execution logic. This biases the design to a specific type of algorithms in the metaheuristic family. Second, given a target problem, most of them design from scratch. Although prior algorithms can be used as starting points, the knowledge regarding what motivates certain design decisions for a problem is untraceable. This knowledge may shed insights into future designs. These two points pose a clear gap towards fully discovering potentials across the metaheuristic family and fertilizing from prior design knowledge.

In this paper, we bridge the gap by proposing the \underline{A}utoregressive \underline{L}earning-based \underline{Des}igner (ALDes) for automated design of metaheuristic algorithms. We provide the following contributions in ALDes:
\begin{itemize}[leftmargin=*]
    \item We formulate metaheuristic algorithm design as a \textit{sequence generation} task. To this we propose a new sequence representation for the algorithm. The representation tokenizes not only algorithmic components and hyperparameters but also ``pointers". Unlike the vector representation within a specific template, the pointers in our representation determine the execution logic of each component, and importantly, can form sequence, branch, and loop executions. This allows the sequence representation to express arbitrary algorithm structures without needing a specific template. 
    \item We harness an \textit{autoregressive} generative network to handle the sequence generation task. Unlike current methods that orthogonally manipulate each entity of a fixed-length vector representation, we borrow the sequential generation ability from the transformer to autoregressively generate entities one at a time, conditioned on previously generated ones. The autoregressive sequence generation pushes the boundary of automated algorithm design to a new level: designing algorithms with diverse lengths and structures, enabling to fully discover potentials over the metaheuristic family. 
    \item We learn from prior design experience by memorizing and retrieving prior knowledge within neurons of the proposed ALDes. Unlike current methods that handle each algorithm design task independently from scratch, we learn a factored embedding of a target problem and learn to generate algorithms upon the problem in an end-to-end fashion. Design knowledge accumulated in neurons of ALDes can be retrieved for algorithm design of new problems, by feeding the new problems. This paves the way to continual design of algorithms for open-ended problem-solving. 
    \item We extensively examine ALDes on 23 pseudo-Boolean optimization (PBO) benchmarks and two real-world applications. Results reveal that comparison baselines with a fixed algorithm structure are not eligible for all problems. ALDes, in contrast, discovers algorithms with different neighborhood definitions, search behaviors, and structures, reasonably fitting for different problems and outperforming all human-created baselines on 24 out of the 25 problems. Further experiments on continual algorithm design tasks illustrate ALDes's ability to leverage prior knowledge for future designs.
\end{itemize}

The remainder of the paper contains: background and related work of automated design of metaheuristic algorithms in Section \ref{sec_literature}, details of the proposed ALDes in Section \ref{sec_designer}, experiments and applications in Section \ref{sec_experiment}, and conclusions in Section \ref{sec_conclusion}.  

\section{Background and Related Work}\label{sec_literature}
\subsection{Automated Design of Metaheuristic Algorithms}
Without loss of generality, the offline automated design of metaheuristic algorithms aims to handle the following task \cite{zhao2023survey}:
\begin{equation}\label{eq_obj}
    \begin{aligned}
        \mathop{\arg\max}\limits_{A\in\mathcal{A}} \ \mathbb{E}_{\mathcal{I}} &\Big[\mathbb{E}_{\mathcal{G}}\big[g(A|i)\big]\Big], \\
        g\in\mathcal{G}, \ i\in\mathcal{I}_t, \ &t=1,2,\cdots,T,
    \end{aligned}
\end{equation}
where $A$ refers to the algorithm to be designed; $\mathcal{A}$ is the design space constituting with algorithmic components and hyperparameters; $g:\mathcal{A}\times\mathcal{I}\to\mathbb{R}$ is $A$'s performance obtained by a run of $A$ on problem instance $i$. The performance can be solution quality, runtime, or anytime performance \textit{etc} \cite{zhao2023survey}. 

Instance $i$ is drawn from the target problem domain $\mathcal{I}_t$, where $t$ refers to the $t$th algorithm design task. In particular, $T=1$ indicates a \textit{one-off} algorithm design task with a fixed target problem domain. The design needs to operate from scratch without prior experience to learn and use. $T>1$ indicates \textit{continual} algorithm design tasks with a series of (or open-ended) target problem domains. The design is allowed to accumulate knowledge from prior tasks, so as to fast adapt to new tasks. This case aligns with the pursuit of general and open-ended (lifelong) automated systems. 

Since metaheuristic algorithms perform stochastic search, one needs to estimate the expected performance over $\mathcal{G}$; in practice this is done by multiple runs of $A$, resulting in multiple values of $g\in\mathcal{G}$. Since the distribution of $i$ over $\mathcal{I}_t$ is unknown, $i$ is usually drawn from a finite set of instances $\mathcal{I}_t^{train}\in\mathcal{I}_t$. Another finite set of instances $\mathcal{I}_t^{test}\in\mathcal{I}_t\backslash\mathcal{I}_t^{train}$ is used to test the designed algorithms.

For continuous problems, algorithmic components in the design space $\mathcal{A}$ can either be computational primitives, \eg, $+, -, *, /$, or existing operators, \eg, Gaussian mutation \cite{back1993overview}. For discrete problems, there is not a clear distinction between primitives and operators, \eg, $swap$ is a representative search operator and also a computational primitive. The design space with operators can be formed with good design choices from prior algorithms and users' domain expertise, thus has got strong track record in many work \cite{bezerra2015automatic,blot2019automatic,bezerra2020automatically,villalon2021pso,yi2022automated,tian2022deep}. In this paper, we follow them and conduct automated design on the \textit{operator} level.  

Methods for offline automated design of metaheuristic algorithms mostly stem from the hyperparameter configuration field. They represent algorithmic components as categorical parameters; the categorical parameters and their conditional hyperparameters are concatenated to form a vector representation of the algorithm; an algorithm can be generated by manipulating the vector representation. Due to the black-box and NP-hard nature of the algorithm design task in Equation \eqref{eq_obj}, search-based methods are dominant in manipulating the vector representation of algorithms. For example, SMAC \cite{hutter2011sequential,lindauer2022smac3} and MIP-EGO \cite{ye2022automated} use Bayesian optimization to build a surrogate function that models the mapping from algorithms to their performance on a target problem. irace \cite{lopez2016irace} also employs model-based search. In particular, it proposes an iterative racing scheme. At each iteration, it first uses racing to select desired algorithms; then the desired algorithms are used to approximate the sampling distribution of algorithms; after that new algorithms are sampled from the distribution for the next iteration. Univariate norm distribution and discrete distribution are the default for numeral and categorical parameters in the vector representation, respectively, in these model-based methods \cite{hutter2011sequential,lopez2016irace}. ParamILS \cite{hutter2009paramils} adopts the model-free iterative local search to exploit desired algorithms; the intensification and capping strategies are proposed to save algorithm evaluations. 

These design methods manipulate a fixed-length vector representation regulated by a specific algorithm template. The template determines feasible execution logic of algorithmic components involved in the representation. For example, simulated annealing (SA) \cite{kirkpatrick1983optimization}, particle swarm optimization (PSO) \cite{shi1998modified}, and genetic algorithm (GA) \cite{holland1973genetic} templates were used to design new variants of SA \cite{franzin2019revisiting}, PSO \cite{villalon2021pso}, and GA \cite{bezerra2015automatic,bezerra2020automatically,yi2022automated}, respectively. This limits to designing a certain type of metaheuristics rather than exploring the whole metaheuristic family. Furthermore, these search-based methods handle each algorithm design task independently from scratch. Although prior algorithms can be used as starting design points, the knowledge regarding what motivates certain design decisions for certain target problems is untraceable, losing insights and principles for future reuse. In this paper, we dedicate to breaking through the two fundamental limitations in offline automated design by proposing the autoregressive learning-based metaheuristic algorithm designer. 

We note that reinforcement learning is another stream of design methods, but they are employed in online automated design in the literature \cite{sakurai2010method,buzdalova2014selecting,adriaensen2016towards,meng2021automated,yi2022automated,yi2023automated}. Using terminologies in reinforcement learning, the state often refers to the information from evaluating the performance of the designed algorithm in the current problem-solving, \eg, features of the target problem instances and features of the algorithm's solutions to the instances; the action is adjusting certain components of the current algorithm; a policy is learned to generate an algorithm with a certain component composition to fit for the problem-solving state. Tabular reinforcement learning, as well as the modern deep Q-network \cite{mnih2015human} and proximal policy optimization \cite{schulman2017proximal}, have been used to learn the policy \cite{sakurai2010method,buzdalova2014selecting,yi2022automated,yi2023automated}. The policy results in an algorithm with adjustable components according to the search trajectory during problem-solving. The structure of the designed algorithm is fixed.

\subsection{Sequence Generation with Transformer} \label{sec_trm}
Transformer is a neural network architecture originally proposed for sequence-to-sequence learning in natural language processing \cite{vaswani2017attention}. Self-attention \cite{parikh2016decomposable,lin2017structured} is the core module of transformer. It learns valuable features from the input, by capturing pairwise relations between tokens in the input sequence \footnote{Tokens in the sequence should first go through an embedding process, if they are originally unstructured. Each token should be combined with a positional encoding, because the tokens in the sequence are ordered but the attention process is invariant to the order.}. Specifically, it first linearly maps the input sequence $\mathbf{X}\in\mathbb{R}^{N\times d}$ ($N$ tokens, each with a size of $d$) to a query, key, and value matrices $\mathbf{Q}$, $\mathbf{K}$, and $\mathbf{V}$, respectively
\begin{equation}\label{eq_trm1}
    \mathbf{Q}=\mathbf{X} \times \mathbf{W}^{q}, \ \mathbf{K}=\mathbf{X} \times \mathbf{W}^{k}, \ \mathbf{V}=\mathbf{X} \times \mathbf{W}^{v},
\end{equation}
where $\mathbf{W}^{q},\mathbf{W}^{k},\mathbf{W}^{v}\in\mathbb{R}^{d\times d^{'}}$ are learnable parameters that augment the features of the input sequence. Then the pairwise relations are captured by 
\begin{equation}\label{eq_trm2}
    \mathbf{X}^{'}={\rm softmax}\Big(\frac{\mathbf{Q} \times \mathbf{K}^\top}{\sqrt{d^{'}}}\Big)\times\mathbf{V},
\end{equation}
where the dot-product score function $\mathbf{Q} \times \mathbf{K}^\top$ scores the pairwise similarity between tokens; $\sqrt{d^{'}}$ scales down the dot productions to stabilize the training; the alignment function ${\rm softmax}(\cdot)$ normalizes the scores; the pairwise relations are calculated by weighted sum of $\mathbf{V}$ with the normalized attention scores.

The self-attention can be conducted in $S$-head subspaces. That is, $\mathbf{X}$ is mapped with $\{\mathbf{W}_{s}^{q},\mathbf{W}_{s}^{k},\mathbf{W}_{s}^{v}\}_{s=1}^{S}$, respectively. Through Equations \eqref{eq_trm1} and \eqref{eq_trm2}, $\{\mathbf{X}_{s}^{'}\}_{s=1}^{S}\in\mathbb{R}^{N\times \frac{d^{'}}{S}}$ are obtained. These $\mathbf{X}_{s}^{'}$ are concatenated to form $\mathbf{X}^{'}$ via ${\rm concat}(\mathbf{X}_{1}^{'},\mathbf{X}_{2}^{'},\cdots,\mathbf{X}_{S}^{'}) \times \mathbf{W}^{0}$, where $\mathbf{W}^{0}\in\mathbb{R}^{d^{'}\times d}$ are learnable parameters. The residual connection and layer normalization, i.e., ${\rm layer\_norm}(\mathbf{X}+\mathbf{X}^{'})$, are then applied to promote information flow. 

The vanilla transformer includes an encoder and a decoder. The encoder uses self-attention to learn valuable features from the input. The decoder uses self-attention to learn relations between tokens in the previously generated sequence segment, in turn to condition the autoregressive generation of the next token. Multiple encoder and decoder blocks can be stacked, respectively, to learn deep features. Please refer to \cite{vaswani2017attention} for the complete transformer architecture.

As an alternative to recurrent neural networks \cite{goodfellow2016deep}, transformer offers strong sequential generation ability by using self-attention to learn through many shortcut relation paths between each pair of tokens in the sequence. This sequential generation ability is appealing for design tasks, e.g., designing neural architectures \cite{zoph2016neural,zoph2018learning}, drug molecules \cite{popova2018deep}, and radiofrequency waveforms \cite{shin2021deep}, since the objects (sequences) to be designed are usually in various structures (lengths), and components of the object (tokens of the sequence) are dependent. The multitude of shortcut relation paths also benefit learning valuable features from complicated inputs (environment of the design tasks), which is significant to partial observable or black-box design environment \cite{loynd2020working}. In this paper, we harness transformer's generation and learning ability to automated design of metaheuristic algorithms.

\section{Proposed ALDes}\label{sec_designer}
Current automated design methodologies generate algorithms within a fixed structure, blocking potentially promising algorithms beyond that structure. Furthermore, they usually operate from scratch. Prior knowledge regarding what motivates certain design decisions for a problem is untraceable, losing insights and principles for future designs. 

We propose the autoregressive learning-based designer, ALDes, to address these issues. Our idea is to formulate metaheuristic algorithm design as a sequence generation task. By \textit{autoregressive} learning to manipulate the \textit{sequence} representation of algorithms, we are able to generate algorithms in various lengths and structures. Furthermore, we learn a factored embedding of a target problem and generate algorithms upon the problem in an end-to-end fashion. This allows prior design knowledge to be traceable for boosting future designs, by feeding the future problem. 

We give the new formulation in Section \ref{sec_formulation}, then detail the workflow of ALDes in \ref{sec_workflow}, and describe how to learn from prior experience for open-ended problem-solving in \ref{sec_open-ended}.

\subsection{Formulation} \label{sec_formulation}
\subsubsection{Algorithm Design as Sequence Generation} 
We formulate the automated design of metaheuristic algorithms as an autoregressive sequence generation task: 
\begin{equation}\label{eq_seq}
    \begin{aligned}
        \mathop{\max}\limits_{\Theta}p_{\Theta}(A)= & \sum_{n=1}^{N}\log p_{\Theta}(a_n|a_{1:(n-1)}), \\
        {\rm s.t.} \ A=\mathop{\arg\max}\limits_{A\in\mathcal{A}}\mathbb{E}_{\mathcal{I}} & \Big[\mathbb{E}_{\mathcal{G}}\big[g(A|i)\big]\Big], \ i\in\mathcal{I}, g\in\mathcal{G},
    \end{aligned}
\end{equation}
where $\Theta$ refers to the parameters of the autoregressive model; $A=(a_{1},a_{2},\cdots,a_N)$ is the sequence representation of the algorithm to be designed; $a_n, n=1,2,\cdots,N$, is a token of the sequence; each token's probability $p$ is conditioned on previous tokens; the joint probability is factorized as a product of the conditionals. Notations in the constraint are the same as those in Equation \eqref{eq_obj}.

We present two appealing features with the new formulation. First, we autoregressively generate entities (tokens) one at a time, conditioned on previously generated ones. This is aligned with the nature that components of a metaheuristic algorithm are dependent and the components may have conditional hyperparameters\footnote{For example, a neighborhood search component has hyperparameters determining the neighborhood region, the uniform mutation component has a hyperparameter determining the mutation probability.}. Second, we can generate sequences in various lengths, enabling us to generate diversified algorithms. In comparison, previous design often orthogonally manipulates each entity of a fixed-length algorithm, falling to capture the dependence and provide diversification.

\subsubsection{Representing Algorithm as a Sequence} \label{sec_sequence}
We propose a new sequence representation of metaheuristic algorithms to fit for the formulation. The sequence representation tokenizes not only algorithmic components and hyperparameters but also \textit{pointers} and \textit{conditions}. Specifically, we treat each component, hyperparameter, pointer, and condition as a token of the sequence, respectively. A component token is followed by its hyperparameter (if any) tokens, a pointer token, and a condition token. They form a snippet. A number of snippets are arranged to form the sequence:
\begin{equation}\label{eq_seq2}
    \begin{aligned}
        A=(S_1,&S_2,\cdots), \\
        {\rm with} \ S_{l}=(a^{{\rm comp}},a^{{\rm param}_{1}},&\cdots,a^{{\rm param}_{C}},a^{{\rm ptr}},a^{{\rm cond}}), \\
        l=1,&2,\cdots,
    \end{aligned}    
\end{equation}
where $A$ is the sequence, $S_l$ is the $l$th snippet, each $a$ is a token, and $C$ is the number of hyperparameters of component $a^{\mathrm{comp}}$.

\begin{figure}[t] 
	\centering
	\subfigure[Tabu search]{\includegraphics[width=1\columnwidth]{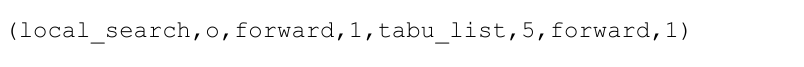}}
        \subfigure[Iterative local search]{\includegraphics[width=1\columnwidth]{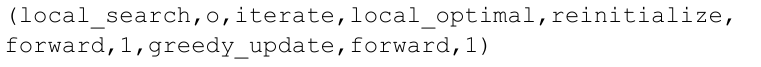}}
        \subfigure[Genetic algorithm]{\includegraphics[width=1\columnwidth]{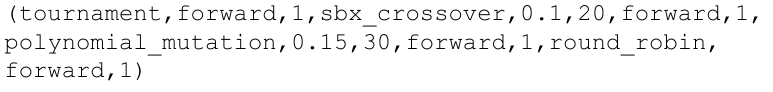}}
	\caption{Three instantiations of the sequence representation. The (a) Tabu search has two components, \texttt{local\_search} and \texttt{tabu\_list}. The \texttt{local\_search} (with hyperparameter \texttt{o} defines the neighborhood region) executes once (count condition \texttt{1}), then the flow \texttt{forward} to update the \texttt{tabu\_list} (hyperparameter \texttt{5} refers the list's length) once (count condition \texttt{1}), after that the flow \texttt{forward} to the next algorithm round. Likewise, in (b), the \texttt{local\_search} iterates until reaching a local optimal (event condition \texttt{local\_optimal}). In (c), \texttt{0.1,20} (\texttt{0.15,30}) are the crossover (mutation) probability and distribution, respectively.}
	\label{fig_seq}
\end{figure}

In a snippet, pointer $a^{\mathrm{ptr}}$ determines the execution flow after running component $a^{\mathrm{comp}}$. We define three pointers, \texttt{forward}, \texttt{iterate}, and \texttt{fork}. The first makes the execution flow head to the component of the next snippet; the middle makes $a^{\mathrm{comp}}$ iterate; the last forks the flow head to the next component and certain subsequent component(s). \texttt{fork} will bring a control parameter token following it, to determine which subsequent components to link to. Condition $a^{\mathrm{cond}}$ regulates the computational cost of $a^{\mathrm{comp}}$. We define two types of conditions, \ie, count-conditions (the number of execution times) and event-conditions (executing until an event happened). Our new representation can form sequence, branch, and loop executions, thereby expressing various algorithm structures without a template. This cannot be done by the common practice of fixed-length vector representation. Figure \ref{fig_seq} shows instantiations of our sequence representation. 

\subsection{Workflow} \label{sec_workflow}
\begin{figure*}[h] 
	\centering
	\includegraphics[width=1.5\columnwidth]{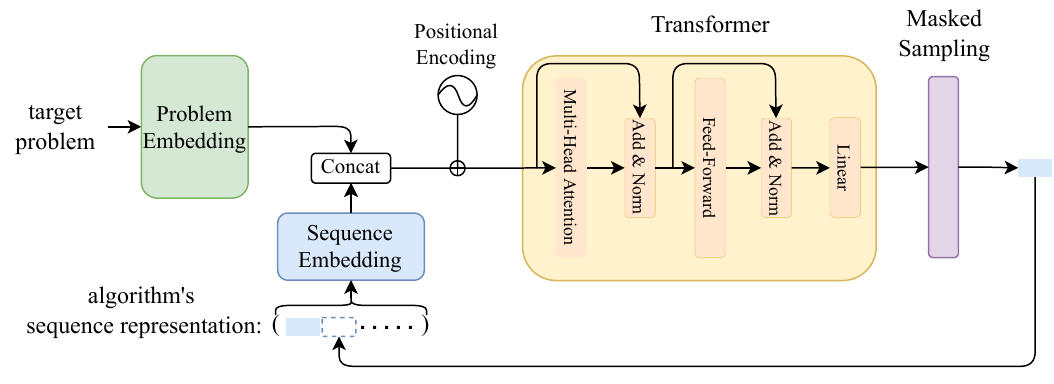}
	\caption{Workflow of the proposed ALDes.}
	\label{fig_designer}
\end{figure*}

We harness an autoregressive generative network to handle the sequence generation task of Equation \eqref{eq_seq}. Figure \ref{fig_designer} shows the workflow, detailed next.

\subsubsection{Architecture}
From Figure \ref{fig_designer}, the network contains four main modules, \ie, problem embedding, sequence embedding, transformer, and masked sampling. We first introduce these modules, respectively, then illustrate the workflow.

Problem embedding embeds the unstructured (\eg, symbolic expression) target problem into ALDes. We elaborate this module in Section \ref{sec_open-ended}, as it is only available for learning from prior experience in \textit{continual} algorithm design tasks (\ie, $T>1$ in Equation \eqref{eq_obj}) and is unnecessary for \textit{one-off} tasks. In continual tasks, ALDes must recognize the target problems in different tasks by embeddings. Thus, it can learn from the design experience on prior problems and extract relevant experience for the design on new problems. While in one-off algorithm design tasks, it does not, as there is only a single fixed target problem domain.

Sequence embedding embeds the input sequence ($A$ in Equation \eqref{eq_seq2}) into ALDes. First, we convert the sequence's tokens to the one-hot format, each is of dimension $M$, where $M$ is the number of all candidate tokens, including algorithmic components, hyperparameters, pointers, conditions, and an `\textit{end}' token that indicates the termination of sequence generation. To enable one-hot tokens, continuous hyperparameters are discretized; while components, pointers, and conditions are naturally discrete. Then, following the common practice of sequence transduction, we learn a linear transformation $\textbf{W}^{seq}\in\mathbb{R}^{M\times d}$ to embed the one-hot tokens. Afterward, the tokens go through sinusoidal positional encoding \cite{vaswani2017attention} to enable position information. 

The transformer module learns features and captures dependencies over tokens of the current sequence, predicting the next token. We use the transformer due to its well-established track records in sequence generation, by self-attention to learn from many shortcut relation paths between tokens of the sequence. Specifically, the Multi-Head Attention and Add \& Norm are as illustrated in Section \ref{sec_trm}. Feed-Forward is a common three-layer perceptron with ReLU activations. Linear is a learnable linear transformation $\textbf{W}^{l}\in\mathbb{R}^{d\times M}$ applied to the last token. The output of the transformer is $\textbf{y}\in\mathbb{R}^M$. 

Masked sampling predicts the next token by sampling over candidate ones. Usual sequence generation tasks involve homogeneous tokens (\eg, words in natural language sequences); each of the tokens should be considered in sampling. Our task, in contrast, involves heterogeneous tokens with specific orders in the sequence (Equation \eqref{eq_seq2}). Given this, we can mask unexpected types of tokens in sampling: All tokens but components are masked at the first round of token prediction. This ensures the sequence starts from a component. Only hyperparameter tokens are unmasked at subsequent rounds, if the component has hyperparameters. Only pointer tokens are unmasked, if the component is hyperparameter-free or the hyperparameters have been generated. Likewise, only condition tokens are unmasked after a pointer token. Only component tokens and the end token are unmasked after a condition token. These principles determine an $M$-dimensional vector mask $\textbf{z}$ with $1$ for unmasked tokens and $-\infty$ for masked tokens. The sampling follows a discrete distribution given by ${\rm softmax}(\textbf{y}\circ\textbf{z})$, where $\circ$ is the Hadamard product. The masked sampling greatly narrows the sampling space, thus easing the sequence generation task.   

Given the above modules, the workflow starts from inputting the algorithm's sequence representation to sequence embedding. The initial sequence is made up of a \textit{`begin'} token, which prompts ALDes to begin the token prediction. After embedding, the sequence goes through positional encoding, transformer, and masked sampling, outputting the next token. The outputted token is then put in the sequence, starting the next round of token prediction. The workflow terminates once the \textit{`end'} token is outputted.  
 
\subsubsection{Training}
Given Equation \eqref{eq_seq}, we train ALDes to maximize the probability of generating the algorithm with the maximum expected performance on the target problem:
\begin{equation} \label{eq_train1}
    J(\Theta)=\mathbb{E}_{p_{\Theta}(a_{1:N})}[R],
\end{equation}
where $\Theta$ refers to learnable parameters, including weights of the linear transformation in sequence embedding ($\textbf{W}^{seq}$) and parameters in the transformer ($\mathbf{W}^q$, $\mathbf{W}^k$, $\mathbf{W}^v$ in Equation \eqref{eq_trm1} and $\textbf{W}^l$); $R=\mathbb{E}_{\mathcal{I}}\big[\mathbb{E}_{\mathcal{G}}[g(A|i)]\big]$, $i\in\mathcal{I}$, $g\in\mathcal{G}$, $A=(a_{1:N})$, is the expected performance, with the same notations as those in Equation \eqref{eq_obj}.

Since $R$ is non-differentiable, we resort to a policy gradient method to iteratively update $\Theta$. Specifically, we use the proximal policy optimization \cite{schulman2017proximal}:
\begin{equation}\label{eq_train2}
    \begin{aligned}
        J(\Theta)=&\mathbb{E}_{k}\Big[\mathop{\min}\big(h_k(\Theta)\hat{R_k},{\rm clip}(h_k(\Theta),1-\epsilon,1+\epsilon)\hat{R_k}\big)\Big], \\
                  &{\rm with} \ h_k(\Theta)=\frac{\prod_{n=1}^{N}p_{\Theta}(a_n^k|a_{1:(n-1)}^k)}{\prod_{n=1}^{N}p_{\Theta_{old}}(a_n^k|a_{1:(n-1)}^k)},
    \end{aligned}
\end{equation}
where $k$ is the index of the $k$th sequence (\ie, algorithm) in a batch; $h_k(\Theta)$ is the weight of importance sampling \cite{schulman2015trust}; $\Theta_{old}$ refers to the parameters before update; $a_n^k$ is the $n$th token of algorithm $k$; $\hat{R_k}=R_k-b$, in which $R_k$ is the $k$th algorithm's expected performance, $b$ is a weighted moving average baseline that reduces the variance of gradient estimate; $\epsilon$ is the clip threshold. The proximal policy optimization allows multiple epochs of minibatch updates, greatly enhancing the training effectiveness concerning standard policy gradient.  

\subsection{Learning From Prior Experience} \label{sec_open-ended}
Unlike search-based design methods that handle each algorithm design task independently from scratch, ALDes can learn and accumulate design knowledge over time in \textit{continual} scenarios to fast adapt to new problems. This aligns with the pursuit of general and open-ended automated systems. To this, we (i) embed the target problems to let ALDes recognize them, such that ALDes can retrieve relevant prior knowledge for new problems; (ii) regularize ALDes's training to accumulate prior knowledge while adapting to new problems. 

For point (i), we use the problem embedding module (Figure \ref{fig_designer}). Intuitively, the embedding might be done by representation learning, \eg, word2vec \cite{mikolov2013efficient} to embed the textual description of the target problem, mathematical language processing \cite{scarlatos2023tree} to embed the symbolic expression of the target problem. But these methods are incomplete, because hard problems are non-trivial to be described by text, symbolic expression is multimodal\footnote{A problem can have different symbolic formulations.}, and many problems do not have symbolic expressions (\eg, black-box ones). 

Given this, we choose a factored embedding, \ie, converting the target problem to a factor representation. The representation is a vector of factors describing the problem's landscape features. The factors are calculated by exploratory landscape analysis \cite{mersmann2011exploratory} in two steps: (1) Sampling the problem's solution space and evaluating the sampled solutions' fitness. The Sobol \cite{sobol1967distribution} and random walk samplings are used for continuous and discrete problems, respectively, following \cite{renau2020exploratory,malan2013survey}. (2) Calculating landscape features with the samples. The features are chosen following \cite{renau2020exploratory} (detailed in the experiment). These features are computationally cheap and are independent of sampling methods. Note that the sampling does \textit{not} incur additional computational overhead, as we randomly select solutions from the sampled ones and reuse them as initial solutions of the generated algorithm $A$, when estimating $A$'s expected performance $R$ (Equation \eqref{eq_train1}). Furthermore, since we use the features as factors to recognize problems rather than analyze the landscape, the choose of features matters \textit{little}. 

The factor representation goes through a learnable linear transformation $\mathbf{W}^{probl}\in\mathbb{R}^{U\times d}$, where $U$ is the number of factors. Afterwards, it serves as a token and concatenates with tokens of the embedded sequence (Figure \ref{fig_designer}). The subsequent transformer attends the factored representation of the target problem and biases the algorithm design upon the problem. The factored embedding does not rely on any specific form of the input target problem, \eg, textual or symbolic expression, thus is a general problem embedding for algorithm design.

For point (ii), we resort to the elastic weight consolidation (EWC) \cite{kirkpatrick2017overcoming}. It can estimate the importance of ALDes's each parameter to accumulating prior design knowledge, and slow down updating the important parameters to adapt to new tasks, so as to balance the accumulation and adaption. Accordingly, the training of ALDes is regularized by
\begin{equation} \label{eq_train3}
    J^{'}(\Theta)=J_{t}(\Theta)+\sum_{r=1}^{R}\frac{\lambda}{2}F_{r}\|\Theta_r-\Theta_{r,t-1}^{*}\|_2^2,
\end{equation}
where $\Theta$ refers to learnable parameters including these from Equation \eqref{eq_train1} plus $\mathbf{W}^{probl}$; $J_{t}(\Theta)$ is the loss of the current algorithm design task and is calculated by Equation \eqref{eq_train2}; $F_r$ is the Fisher information matrix estimating the importance of the $r$th parameter to memorizing prior knowledge; $\Theta_{r,t-1}^{*}$ is the $r$th (learned) parameter for previous tasks; $\lambda$ sets the weight. We resort to this well-established regularization method, because it is scalable and will not incur overload in terms of memory and model scale, concerning replay and parameter isolation methods \cite{de2021continual}.    
  
\section{Experiments and Applications}\label{sec_experiment}
We do experiments and applications to evaluate (i) the proposed ALDes's efficiency in generating high-performance and diversified metaheuristic algorithms, and the necessity of such diversification to fit for different problem-solving, (ii) the efficiency of ALDes in learning from prior design experience to facilitate future designs.

\subsection{Setup} \label{sec_setup}
\subsubsection{Target problems}
They include the representative PBO numerical benchmarks \cite{DoerrYHWSB20} and two real-world problems. PBO is a well-established discrete optimization problem suite with various fitness landscapes. 23 PBO functions are used, including OneMax, its Harmonic weighted version, W-model extensions (F1, F3, F4-F10), LeadingOnes and its W-model extensions (F2, F11-F17), Low Autocorrelation Binary Sequences (F18), Ising Models (F19-F21), Maximum Independent Vertex Set (F22), and N-Queens (F23). Each function is treated as an independent target problem with four instances in different solution space dimensions: $100$, $225$, $400$, and $625d$. 

The real-world problems are for investigating ALDes's practicality. The first is the discrete, non-separable beamforming problem from a reconfigurable intelligent surface (RIS)-aided communication system. The second is the power system restoration problem with a highly constrained binary solution space. Each problem presents instances with different input data. Metaheuristics could handle the discrete, highly constrained, and non-separable challenges. The automated design could customize efficient algorithms from various options.

\subsubsection{Baselines}
Baselines span various algorithms across the metaheuristic family for a comprehensive comparison. They include iterative local search (ILS) \cite{lourencco2003iterated}, simulated annealing (SA) \cite{kirkpatrick1983optimization}, Tabu search (TS) \cite{glover1989tabu}, and genetic algorithm (GA) \cite{holland1973genetic}. The neighborhood structure of ILS, SA, and TS is a common practice of $1d$ distinct from the current solution; other settings align with established practice of \cite{glover2006handbook}, summarized in Table \ref{tab_settings}. GA uses uniform crossover and standard bit mutation unless stated otherwise, with hyperparameters (crossover probability $\eta_c$ and the number of bits to be flipped $\eta_m$) configured by irace \cite{lopez2016irace} for each target problem, according to settings of \cite{ye2022automated}. The analytical baseline CPLEX is used for real-world problems to access ALDes's practicality.

\subsubsection{Implementation details} \label{sec_implementation}
We implement the proposed ALDes as follows: (i) Size: The output dimension is $32$ (\ie, $d=d^{'}=32$) for all learnable parameters except for $\textbf{W}^l$. $8$ heads are used in the attention. This model structure is straightforward. More complex structures may boost performance but are not the focus of this paper. 

(ii) Training: For each PBO problem, the $100$, $225$, and $400d$ instances are for training; for each real-world problem, half of its instances are for training. On each problem, ALDes is trained for 100 epochs with a batch size of 16 algorithms generated per epoch. Update iteration=$5$ within each proximal policy optimization. For evaluating algorithms' expected performance ($R$ in Equation \eqref{eq_train1}), each generated algorithm runs 5 times per training instance with 5000 function evaluations (FEs) per run. The performance is measured by the best solution fitness, and expected performance is averaged over all runs. More evaluation budget could stabilize training but increase computation load. An algorithm contains up to $6$ components to avoid unnecessary complexity. Adam optimizer \cite{kingma2014adam} with annealing learning rate starting from $5e-5$ is used. $\epsilon=0.2$, $\lambda=200$ following the common practice. The design space in Table \ref{tab_space} is employed, detailed in Appendix \ref{appendix_space}. 

(iii) Testing: The $625d$ instance of each PBO problem and the remaining instances of each real-world problem are used for testing. The algorithm inferred from the trained ALDes is compared with the baselines on test instances. All algorithms perform population-based search (population size=$50$) for fair peer comparison. Each algorithm runs 30 times per problem with 50000 FEs per run.

\subsection{Efficiency in Generating High-Performance and Diversified Algorithms} \label{sec_exp1}
In this subsection, we train the proposed ALDes from scratch on each PBO problem \textit{respectively} to simulate one-off algorithm design tasks. Table \ref{tab_pbo} presents results on \textit{test} problem instances, where $Alg^*$ is the algorithm inferred from the trained ALDes; larger values refer to better performance. 

\begin{table*}[t]
\centering
% \scriptsize
\caption{Performance (mean$\pm$std) over all runs on the test instance of each PBO problem. $Alg^*$'s performance is averaged over 5 algorithms inferred from ALDes's 5 training trials with different seeds. Best results according to pair-wise Wilcoxon sign test ($5\%$ significance level) are with grey background.}
\label{tab_pbo}
\begin{tabular}{lrrrrr}
\toprule
    & \multicolumn{1}{c}{ILS}  & \multicolumn{1}{c}{SA}   & \multicolumn{1}{c}{TS}   & \multicolumn{1}{c}{GA}   & \multicolumn{1}{c}{$Alg^*$}\\
\midrule
F1  & 609.40    $\pm$ 2.87E+00 & 371.90    $\pm$ 3.24E+01 & 394.30    $\pm$ 1.75E+02 & 452.63    $\pm$ 7.96E+00 & \cellcolor{gray}625.00    $\pm$ 0.00E+00 \\
F2  & 93.67     $\pm$ 8.97E+01 & 15.60     $\pm$ 8.94E+00 & \cellcolor{gray}161.90    $\pm$ 2.66E+01 & 54.40     $\pm$ 3.14E+00 & 158.67    $\pm$ 2.29E+01 \\
F3  & 192550.47 $\pm$ 1.13E+05 & 119547.47 $\pm$ 4.26E+06 & 142345.33 $\pm$ 1.77E+07 & 146910.90 $\pm$ 1.27E+06 & \cellcolor{gray}195625.00 $\pm$ 0.00E+00 \\
F4  & \cellcolor{gray}312.00    $\pm$ 0.00E+00 & 201.60    $\pm$ 1.87E+01 & 236.47    $\pm$ 6.59E+01 & 250.83    $\pm$ 3.80E+00 & \cellcolor{gray}312.00    $\pm$ 0.00E+00 \\
F5  & 555.37    $\pm$ 1.96E+00 & 337.10    $\pm$ 1.60E+01 & 361.43    $\pm$ 9.59E+01 & 413.83    $\pm$ 1.17E+01 & \cellcolor{gray}562.00    $\pm$ 0.00E+00 \\
F6  & 203.80    $\pm$ 5.34E+00 & 143.67    $\pm$ 7.82E+00 & 181.13    $\pm$ 1.66E+01 & 170.33    $\pm$ 2.30E+00 & \cellcolor{gray}207.90    $\pm$ 9.31E-02 \\
F7  & 504.17    $\pm$ 3.35E+01 & 380.50    $\pm$ 1.87E+01 & 488.50    $\pm$ 5.93E+01 & 386.30    $\pm$ 2.01E+01 & \cellcolor{gray}510.57    $\pm$ 2.29E+01 \\
F8  & 303.40    $\pm$ 1.28E+00 & 186.43    $\pm$ 8.81E+00 & 156.80    $\pm$ 3.49E+01 & 227.47    $\pm$ 1.77E+00 & \cellcolor{gray}314.00    $\pm$ 0.00E+00 \\
F9  & 591.40    $\pm$ 1.26E+02 & 368.67    $\pm$ 2.91E+01 & 314.93    $\pm$ 1.38E+02 & 452.23    $\pm$ 8.81E+00 & \cellcolor{gray}625.00    $\pm$ 0.00E+00 \\
F10 & 367.00    $\pm$ 3.72E+01 & 368.33    $\pm$ 2.71E+01 & 316.67    $\pm$ 1.26E+02 & 450.87    $\pm$ 1.06E+01 & \cellcolor{gray}598.50    $\pm$ 5.58E+01 \\
F11 & 92.67     $\pm$ 3.05E+02 & 14.63     $\pm$ 1.20E+01 & \cellcolor{gray}160.40    $\pm$ 1.91E+02 & 54.20     $\pm$ 6.58E+00 & \cellcolor{gray}161.87    $\pm$ 5.10E+01 \\
F12 & 93.90     $\pm$ 1.20E+02 & 14.03     $\pm$ 1.43E+01 & \cellcolor{gray}163.23    $\pm$ 1.96E+02 & 55.10     $\pm$ 6.30E+00 & \cellcolor{gray}164.20    $\pm$ 3.95E+01 \\
F13 & 23.57     $\pm$ 9.61E+01 & 17.37     $\pm$ 9.83E+00 & 7.20      $\pm$ 4.06E+01 & 44.07     $\pm$ 3.58E+00 & \cellcolor{gray}89.63     $\pm$ 9.07E+00 \\
F14 & 17.30     $\pm$ 2.04E+01 & 11.97     $\pm$ 1.30E+01 & 4.37      $\pm$ 8.90E+00 & 44.43     $\pm$ 4.74E+01 & \cellcolor{gray}47.57     $\pm$ 4.19E+01 \\
F15 & 12.17     $\pm$ 1.11E+01 & 8.93      $\pm$ 3.58E+00 & 4.60      $\pm$ 6.11E+00 & 28.30     $\pm$ 2.29E+00 & \cellcolor{gray}35.20     $\pm$ 1.08E+01 \\
F16 & 18.27     $\pm$ 3.99E+01 & 13.83     $\pm$ 5.87E+00 & 7.40      $\pm$ 2.94E+01 & 52.73     $\pm$ 6.55E+00 & \cellcolor{gray}67.83     $\pm$ 1.08E+01 \\
F17 & 16.00     $\pm$ 7.93E+00 & 12.87     $\pm$ 9.15E+00 & 5.33      $\pm$ 4.89E+00 & 42.27     $\pm$ 4.30E+01 & \cellcolor{gray}44.27     $\pm$ 1.44E+02 \\
F18 & 3.73      $\pm$ 1.30E-02 & 1.87      $\pm$ 2.18E-03 & 3.73      $\pm$ 1.27E-02 & 1.51      $\pm$ 1.32E-03 & \cellcolor{gray}4.28      $\pm$ 1.43E-02 \\
F19 & 542.60    $\pm$ 6.96E+01 & 376.20    $\pm$ 2.33E+01 & 474.13    $\pm$ 1.68E+02 & 401.13    $\pm$ 8.81E+00 & \cellcolor{gray}593.00    $\pm$ 2.43E+01 \\
F20 & 1052.33   $\pm$ 1.64E+02 & 708.13    $\pm$ 6.07E+01 & 866.53    $\pm$ 1.82E+02 & 762.00    $\pm$ 5.85E+01 & \cellcolor{gray}1110.47   $\pm$ 5.18E+02 \\
F21 & 1581.93   $\pm$ 5.57E+02 & 1039.20   $\pm$ 7.84E+01 & 1289.40   $\pm$ 4.46E+02 & 1129.07   $\pm$ 6.17E+01 & \cellcolor{gray}1724.13   $\pm$ 4.42E+02 \\
F22 & 247.50    $\pm$ 1.78E+01 & -77674.60 $\pm$ 5.44E+07 & -39790.03 $\pm$ 7.27E+07 & -55636.10 $\pm$ 4.60E+06 & \cellcolor{gray}273.67    $\pm$ 2.40E+01 \\
F23 & -216.90   $\pm$ 4.33E+03 & -20644.67 $\pm$ 4.71E+05 & -18885.37 $\pm$ 1.28E+06 & -13670.17 $\pm$ 6.85E+04 & \cellcolor{gray}23.47     $\pm$ 2.57E-01 \\
\bottomrule
\end{tabular}
\end{table*}

\begin{figure}[t] 
	\centering
        \subfigure[F1]{\includegraphics[width=0.46\columnwidth]{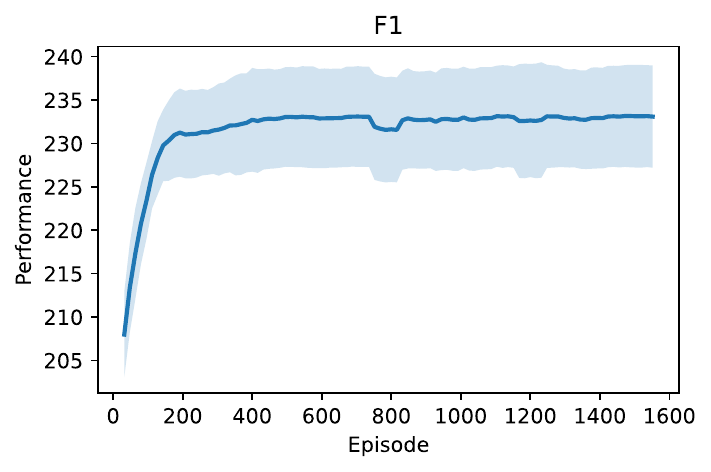}\label{fig_F1}}
        \subfigure[F3]{\includegraphics[width=0.48\columnwidth]{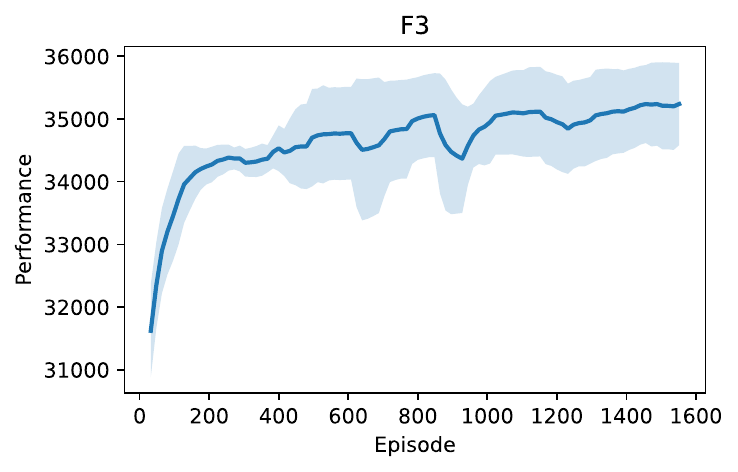}\label{fig_F3}}
        \subfigure[F14]{\includegraphics[width=0.46\columnwidth]{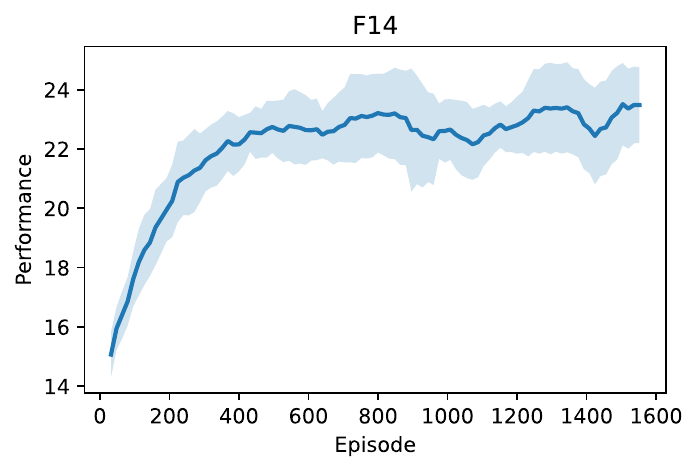}\label{fig_F14}}  
        \subfigure[F15]{\includegraphics[width=0.48\columnwidth]{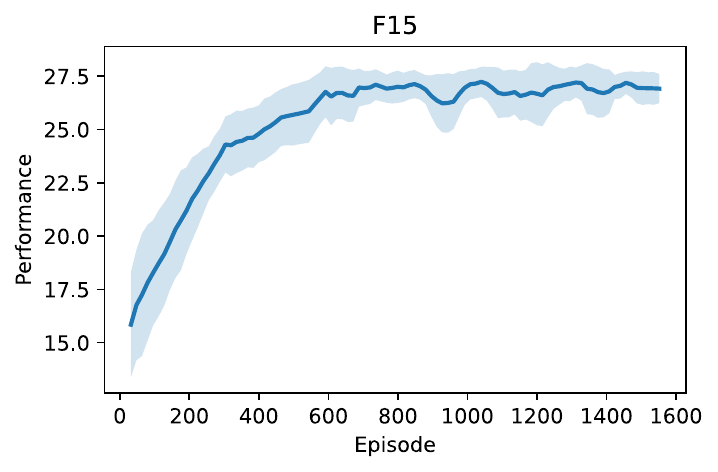}\label{fig_F15}}
        \subfigure[F17]{\includegraphics[width=0.46\columnwidth]{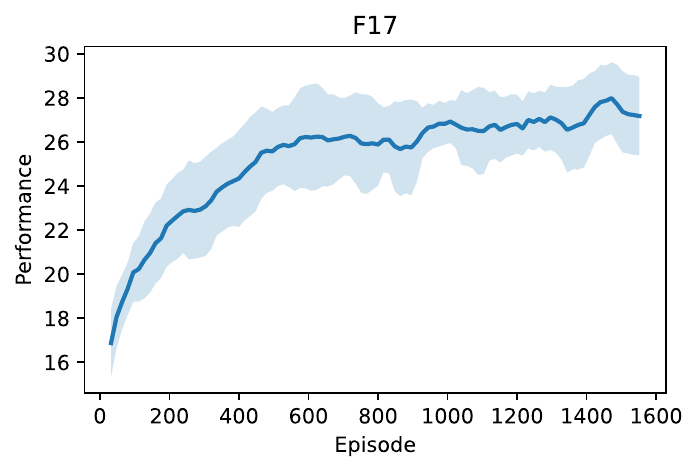}\label{fig_F17}} 
        \subfigure[F20]{\includegraphics[width=0.48\columnwidth]{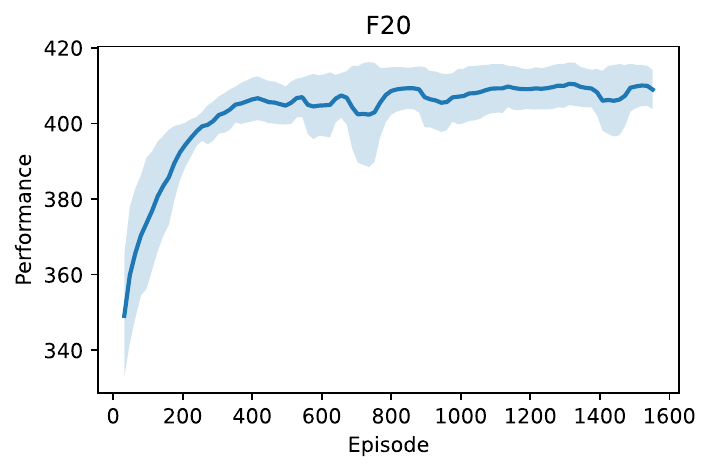}\label{fig_F20}}
	\caption{ALDes's training performance versus episodes. Averaged over 5 trails.}
	\label{fig_one_off}
\end{figure}

F1, F3, and F4-F10 are OneMax, its Harmonic weighted version, and W-model extensions, respectively. OneMax can be solved by greedily flipping each bit from left to right. But the baselines, without considering such ordinal attribute in their neighborhood structure, fail to reach optima in the $625d$ solution space. The algorithm $Alg^*$ generated by ALDes (see Algorithm \ref{alg_f1}\footnote{ALDes is trained 5 trials with different seeds. This results in 5 algorithms by inference from each trial (averaged performance is given in Table \ref{tab_pbo}). We detail the algorithm with the best performance to illustrate ALDes's potential. The same goes for Algorithms \ref{alg_f13}-\ref{alg_restoration}.}) is a variable neighborhood search (VNS) \cite{mladenovic1997variable}-style algorithm utilizing two neighborhood structures: one is $1d$ distinct from the current solution (\texttt{reset\_n}, $n=1$); the other is determined by the uniform crossover of probability $0.7$ and random reset of probability $0.1$. Furthermore, the neighborhood structures are exploited by population-based search (roulette wheel for choosing solutions to search from, and pairwise selection for updating solutions). Such variable neighborhood structures and diversity maintenance ability enable $Alg^*$ reaching optima in all 30 runs within the limited FEs. Figure \ref{fig_F1} shows the training curve. ALDes converges well on this simple problem. Similar results are observed on F3 (Table \ref{tab_pbo}, Figure \ref{fig_F3}), the linear weighted version of F1. On F4-F10, the W-model extensions of F1 with dummy bits, neutrality, epistasis, and fitness perturbation in the landscape, ALDes shows constant outperformance.

\begin{algorithm}[t]
\caption{Pseudocode of the $Alg^*$ for F1} 
\label{alg_f1}
\footnotesize
\KwIn{initial solutions $S$}
\While{algorithm not terminate}{
    \While{\texttt{count\_5\%*FE} \ not \ met}{
    $S=\texttt{roulette\_wheel}(S)$ \\					
    $S_{new}=\texttt{reset\_n}(S,n=1)$ \\
    $S=\texttt{piarwise\_select}(S,S_{new})$ \\
    }
    \While{\texttt{count\_5\%*FE} \ not \ met}{
    $S=\texttt{roulette\_wheel}(S)$ \\					
    $S_{new}=\texttt{cross\_uniform}(S,p=0.7)$ \\
    $S_{new}=\texttt{reset\_rand}(S_{new},p=0.1)$ \\
    $S=\texttt{pairwise\_select}(S,S_{new})$  \\
    }
}
\KwOut{best solution from $S$}
\end{algorithm}

F2 and F11-F17 are LeadingOnes and its W-model extensions. They are more difficult than the OneMax series due to their non-separability, thus the baselines' results dramatically degrade. An exception is TS, which performs well on F2 and dummy bit variants (F11 and F12) but deteriorates on F13-F17. This implies that locating local optima by the Tabu list is insufficient to handle neutrality, epistasis, and fitness perturbation. ALDes overall works better. Specially, we observe interesting results on F13. The algorithm $Alg^*$ generated by ALDes (Algorithms \ref{alg_f13}) is a GA variant that always selects offspring $S_{new}$ for the next generation ($\texttt{alway\_select}$). This unbiased algorithm makes sense to F13 with neutral landscape. The GA baseline is configured with $\eta_c=0.05,\eta_m=0.05$ for F13. This relative steady-state search is also reasonable for neutrality, thus yields the second-best result. On F15, ALDes generates an ILS-style algorithm (Algorithms \ref{alg_f15}), in which the local search of two-point crossover and random reset with probability of $0.1$ iterates $10\%\times$FEs times, follows by global perturbation ($\texttt{reinitialize}$). This ILS variant much outperforms the ILS baseline. The larger magnitude of local search (by crossover and mutation) contributes to this, for the rugged and deceptive landscape. This observation is similar to that in \cite{DoerrYHWSB20}. For training efficiency, results on F14, F15, and F17, the three hardest problems in the LeadingOnes category, are shown in Figures \ref{fig_F14}, \ref{fig_F15}, and \ref{fig_F17}, respectively. Despite fluctuation, ALDes converges on training and performs well on test (Table \ref{tab_pbo}), which confirms the efficiency.

\begin{algorithm}[t]
\caption{Pseudocode of the $Alg^*$ for F13} 
\label{alg_f13}
\footnotesize
\KwIn{initial solutions $S$}
\While{algorithm not terminate}{
    $S=\texttt{roulette\_wheel}(S)$ \\					
    $S_{new}=\texttt{cross\_n}(S,n=2)$ \\
    $S_{new}=\texttt{reset\_rand}(S_{new},p=0.1)$ \\
    $S=\texttt{alway\_select}(S,S_{new})$  \\
}
\KwOut{best solution from $S$}
\end{algorithm}

\begin{algorithm}[t]
\caption{Pseudocode of the $Alg^*$ for F15} 
\label{alg_f15}
\footnotesize
\KwIn{initial solutions $S$}
\While{algorithm not terminate}{
    \While{\texttt{count\_10\%*FE} \ not \ met}{
    $S=\texttt{tournament}(S)$ \\					
    $S_{new}=\texttt{cross\_n}(S,n=2)$ \\
    $S_{new}=\texttt{reset\_n}(S_{new},n=1)$ \\
    $S=\texttt{piarwise\_select}(S,S_{new})$ \\
    }
    $S=\texttt{reinitialize}(S)$ \\
}
\KwOut{best solution from $S$}
\end{algorithm}

F19-F21 are Ising Models. They are NP-hard quadratic Hamiltonian energy functions of $d$ spins up or down ($d$ binary bits) in a lattice \cite{lucas2014ising}. ALDes performs the best on these problems. For illustration, the generated algorithm $Alg^*$ for F20 is given in Algorithm \ref{alg_f20}. It is a VNS-style algorithm with two neighborhood structures. One is a small neighborhood of one bit distinct from the current solution; the other is a large neighborhood determined by the uniform crossover (probability of 0.9) and random reset (probability of 0.1). The neighborhood structures are exploited by population-based search (roulette wheel for choosing solutions to search from, and round robin for updating solutions). In comparison, the competitors with a single search structure are inferior. This suggests that the alternation of small and large neighborhood structures, together with population-based search, benefit the search of highly-coupled spins in Ising models. This observation is similar to that in \cite{chen2022optimizing}, in which a slow quantum adiabatic, scheduled by the global Monte Carlo tree search, got remarkable performance. Eligible training performance is observed in Figure \ref{fig_F20}.

\begin{algorithm}[t]
\caption{Pseudocode of the $Alg^*$ for F20} 
\label{alg_f20}
\footnotesize
\KwIn{initial solutions $S$}
\While{algorithm not terminate}{
    \While{\texttt{count\_10\%*FE} \ not \ met}{
    $S=\texttt{roulette\_wheel}(S)$ \\	
    $S_{new}=\texttt{reset\_n}(S,n=1)$ \\
    $S=\texttt{round\_robin\_select}(S,S_{new})$ \\
    }
    \While{\texttt{count\_10\%*FE} \ not \ met}{
    $S=\texttt{roulette\_wheel}(S)$ \\	
    $S_{new}=\texttt{cross\_uniform}(S,p=0.9)$ \\
    $S_{new}=\texttt{reset\_rand}(S_{new},p=0.1)$ \\
    $S=\texttt{round\_robin\_select}(S,S_{new})$  \\
    }
}
\KwOut{best solution from $S$}
\end{algorithm}

ALDes also achieves leading performance on F18, F22, and F23. On F18, ALDes reaches $4.28$ within $50000$ FEs, which is even better than the target value (4.26) obtained by 12 representative metaheuristic algorithms, each within $5\times d^2$ ($d=625$) FEs, in \cite{DoerrYHWSB20}. On F22, SA, TS, and GA's results are ineligible, implying that they may fall into the strong basin of attraction of this problem. Similar results are observed on F23, in which ALDes reaches 23.47, very close to the target value (24) in \cite{DoerrYHWSB20}.

Put together, we observe three significant findings. First, the comparison on $625d$ test instances, which is of much higher dimension than ALDes's training instances ($100$, $225$, and $400d$), has placed ALDes at a disadvantage. Even so, ALDes wins the best on 22 out of 23 problems, and in most problems, outperforms competitors by a large margin. This clearly confirms ALDes's efficiency in generating high-performance algorithms. Second, the baselines' performance ranks are relatively balanced across the problems. This supports our motivation that a single fixed algorithm structure is not always the best for different problem-solving. Third, ALDes generates diversified algorithms for different problems, \eg, the VNS-, ILS-, and GA-style ones with distinct behaviors (Algorithms \ref{alg_f1}-\ref{alg_f20}). It also generates algorithms beyond intuition, such as the unbiased one for handling neutrality (Algorithm \ref{alg_f13}). The generated algorithms perform reasonably well on corresponding problems, as analyzed above. All these clearly demonstrate ALDes's efficiency in generating diversified algorithms across the metaheuristic family to suit different problem-solving.

\subsection{Efficiency in Learning from Prior Experience} \label{sec_exp2}
In this subsection, we \textit{continually} train the proposed ALDes on a series of PBO problems to simulate continual algorithm design tasks. We construct two continual tasks from the PBO suite: Task 1, comprising 6 OneMax versions (F1-F6) and 1 LeadingOnes (F7), is relatively easy. Task 2, consisting of F1, F2, F11, F18, F19, F22, F23 from different categories (\eg, OneMax, LeadingOnes, Ising, etc), is challenging, which augments concept drift across the problems. For each task, starting from scratch, ALDes is trained on the problems one by one using Equation \eqref{eq_train3} and the same settings as outlined in Section \ref{sec_implementation}. Factors for problem embedding are detailed in Appendix \ref{appendix_problem}.

\begin{figure}[t] 
	\centering
	\subfigure[Test results with EWC on continual task 1.]{\includegraphics[width=1\columnwidth]{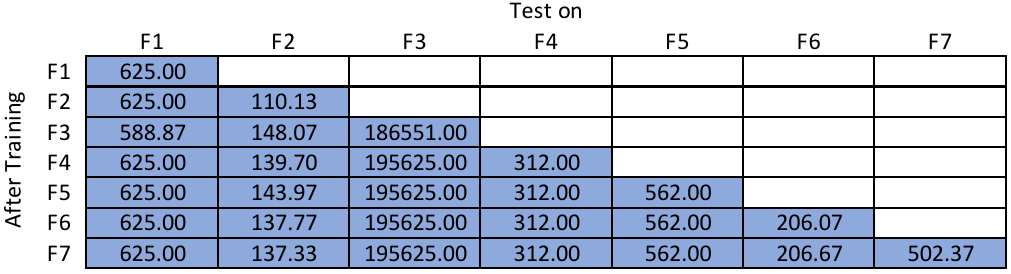}}
        \subfigure[Test results without EWC on continual task 1.]{\includegraphics[width=1\columnwidth]{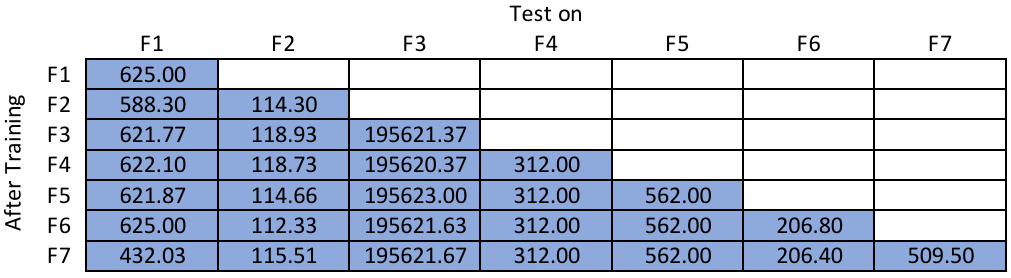}}
        \subfigure[Test results with EWC on continual task 2.]{\includegraphics[width=1\columnwidth]{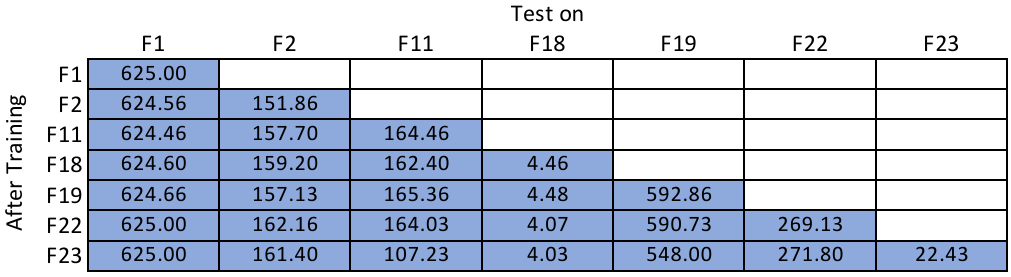}}
        \subfigure[Test results without EWC on continual task 2.]{\includegraphics[width=1\columnwidth]{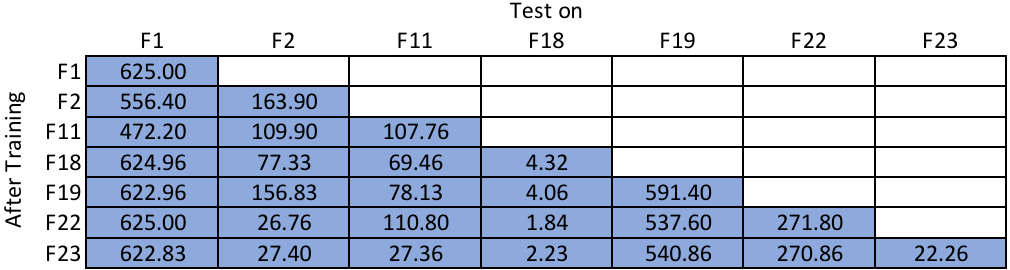}}
	\caption{Average performance over all runs of five generated algorithms on the test instance of each problem. The five algorithms are inferred from ALDes's five training trials with different seeds. In each subfigure, the $i$th row reports the test results after training on the first $i$ problems. For example, the $(3,2)$th cell of subfigure (a) shows the performance of algorithms for F2 ($148.07$), in which algorithms are inferred from the ALDes that has been trained on F1, F2, and F3. }
	\label{fig_ewc}
\end{figure}

To examine ALDes's ability to accumulate prior knowledge, we compare it to the variant without EWC (\ie, train using Equation \eqref{eq_train2}, other settings keep constant). This variant behaves akin to \textit{fine-tuning}, \ie, adapting to new tasks without considering prior ones. Figure \ref{fig_ewc} reports the results. In task 1, with the training progress along the problems, ALDes with EWC constantly maintains its performance on previous problems, while the variant without EWC fails to do so on F1 and F2. This trend is more pronounced in the harder task 2, where the performance without EWC drops by half on F2, F11, and F18 as training progresses. ALDes with EWC performs relatively stable, showing its capability for knowledge accumulation. 

\begin{figure}[t] 
	\centering
        \subfigure[F11]{\includegraphics[width=0.48\columnwidth]{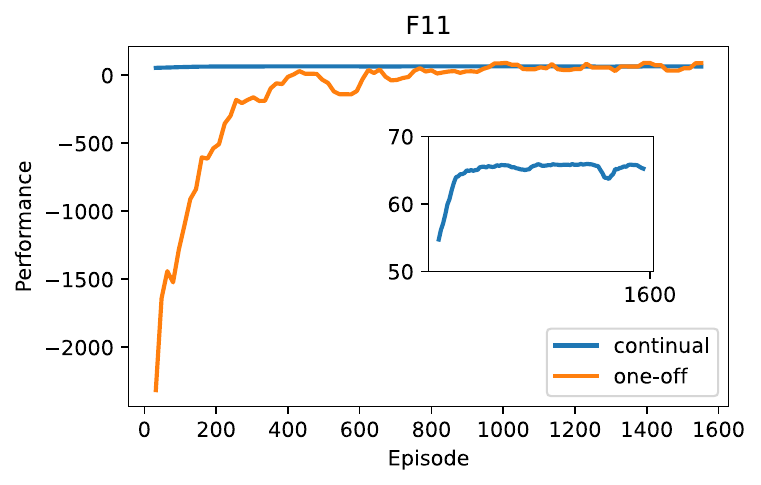}\label{fig_F11}}
        \subfigure[F18]{\includegraphics[width=0.48\columnwidth]{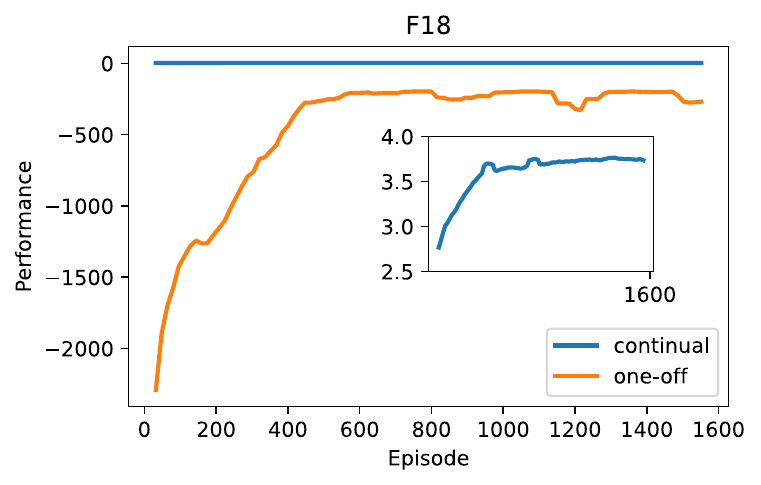}\label{fig_F18}}  
        \subfigure[F19]{\includegraphics[width=0.48\columnwidth]{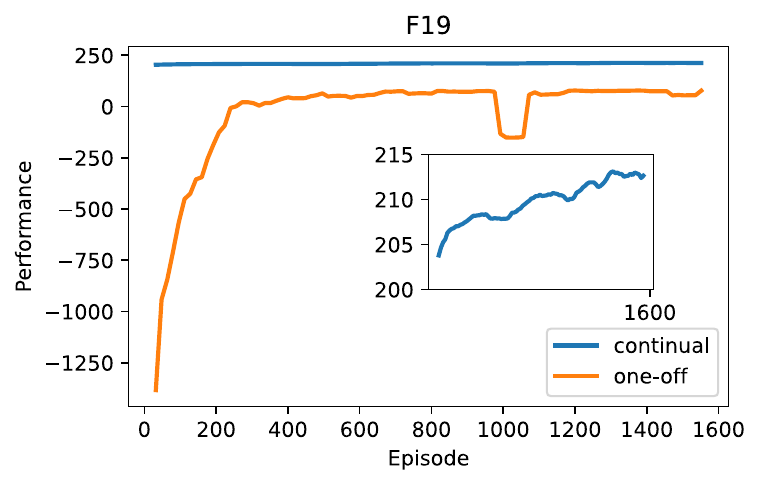}\label{fig_F19}} 
        \subfigure[F22]{\includegraphics[width=0.48\columnwidth]{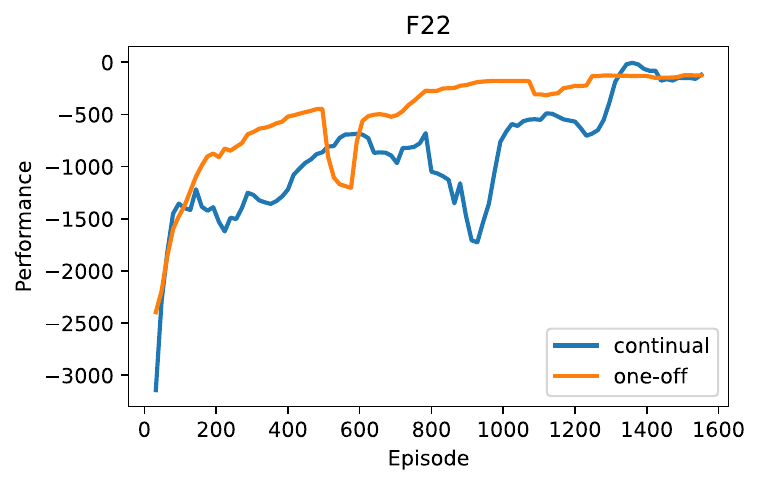}\label{fig_F22}}
	\caption{ALDes's training performance versus episodes. Averaged over 5 trails. The one-off curves report the results of training from scratch. The continual curves report the results of training under the continual task settings. The insets resize the continual curves to show the convergent tendency.}
	\label{fig_continual}
\end{figure}

We further investigate whether the accumulated knowledge can facilitate the design on future problems. Figure \ref{fig_continual} depicts training curves of F11, F18, F19, and F22 in task 2. Continual training on F11, F18, and F19 starts from much higher performance than training from scratch. The accumulated knowledge contributes to this fast adaption. Performance improves further as training progresses (see Insets in Figure \ref{fig_continual}). The adaption is not observed on F22, indicating significant concept drift between its landscape (with a strong basin of attraction) and prior problems. Nonetheless, the training converges to a comparable performance with the one-off curve. Overall, results of Figures \ref{fig_ewc} and \ref{fig_continual} suggest that with EWC, ALDes is able to accumulates knowledge learned from prior tasks. The knowledge enables ALDes to fast adapt to future designs. Even in cases of significant concept drift, ALDes exhibits eligible convergence. All these showcase ALDes's potential for continual design of algorithms for open-ended problem-solving.    

\subsection{Application to Beamforming in RIS-aided Communications} \label{sec_app2}
\subsubsection{Problem Description} 
The reconfigurable intelligent surface (RIS) is a cutting-edge technology designed to improve cost-effective communications \cite{yuan2020intelligent}. Functioning as a planar passive radio structure, it features reconfigurable passive elements that can independently adjust the phase shift of incoming signals. By working together, these elements generate a directional beam, enhancing signal quality. 

Specifically, we consider the RIS-aided downlink multi-user MISO system detailed in \cite{yan2022fitness}. In this setup, a BS with multiple antennas serves K single-antenna users via an RIS with N elements, facilitating non-line-of-sight links. The aim is to maximize the sum rate of all users while adhering to transmit power constraints, achieved through joint optimization of BS beamforming and RIS phase shifts \cite{yan2022fitness}:
\begin{subequations}
	\label{eq-maxSumRate}
	\begin{align}
        &\max_{\wt_k, \mTa}\quad \sum_{k=1}^{K}\log _{2}(1+\frac{|(\bht_{\fd,k}^\mathrm{H}+\bht_{\fr,k}^\mathrm{H}\mTa\Gt)\wt_k|^2}{\sum_{j\neq k}^K| (\bht_{\fd,k}^\mathrm{H}+\bht_{\fr,k}^\mathrm{H}\mTa\Gt)\wt_{j}|^{2}+\sigma ^{2}}),\tag{\ref{eq-maxSumRate}{a}}\label{eq-Pa} \\ 
        &\quad s.t.\quad \theta_n=\beta_ne^{j\phi_n},\tag{\ref{eq-maxSumRate}{b}}\label{eq-Pb} \\ 
        &\qquad\quad\ \phi_n=\frac{\tau_n2\pi}{2^b}, \tau_n\in\{0,...,2^b-1\},\tag{\ref{eq-maxSumRate}{c}}\label{eq-Pc} \\
        &\qquad\quad\ \sum_{k=1}^{K}\|\mathbf{w}_k\|^{2}\leq P_{T},\tag{\ref{eq-maxSumRate}{d}}\label{eq-Pd}		
	\end{align}
\end{subequations}
where $\wt_k\in\bbC^{M\times 1}$ is the active beamforming at the BS towards user $k$; $\mTa=diag(\theta_1,...,\theta_n,...,\theta_N)$ is a diagonal matrix with RIS phase-shifts being the diagonal values; \eqref{eq-Pc} represents that RIS phase shifts take $b$-bit values; $\bht_{\fd,k}\in\bbC^{M\times1}$, $\Gt\in\bbC^{N\times M}$, and $\bht_{\fr,k}\in\bbC^{N\times1}$ are the channels BS-user $k$, BS-RIS, and RIS-user $k$, respectively, which are modelled as random matrices; $\beta_n=1$ is for all RIS elements; \eqref{eq-Pd} restricts the transmit power being not larger than $P_T$. 

The problem is non-convex and NP-hard. Additionally, landscape analysis indicates rugged and scattered local optimums, particularly with large-scale RIS elements \cite{yan2022fitness}. The typical solver employs water-filling \cite{yu2004iterative} for BS beamforming and individual estimation of RIS element phase shifts. However, decoupled estimation is deemed inadequate \cite{yan2022fitness}. Metaheuristics offer potential for globally searching the rugged and highly coupled space.

\subsubsection{Applying ALDes} 
We apply the proposed ALDes to generate metaheuristic algorithms for estimating RIS phase shifts. Target problem instances differ in the number of RIS elements. Baselines include random beamforming, sequential beamforming \cite{di2020hybrid}, ILS, SA, and GA \footnote{Sequential beamforming exhaustively adjusts the phase shift of each element individually while maintaining the remaining phase shifts unchanged. GA consists of tournament mating selection, one-point crossover, random mutation of probability $0.2$, and round-robin environmental selection, following \cite{yan2022fitness}. ILS, SA, and other settings are the same as Section \ref{sec_setup}.}. Table \ref{tab_beam} reports the results on test instances, in which $Alg_{\rm beam}^*$ is the algorithm generated by ALDes. All metaheuristic peers handle the constraints by penalizing solutions' fitness with constraint violations. The results measure the quality of service for all users; smaller values refer to better service quality.

$Alg_{\rm beam}^*$ obtains superior performance on all instances. It is a GA-style algorithm (Algorithm \ref{alg_beam}). Its difference from the baseline GA is the $1d$ reset and greedy selection. The difference enables $Alg_{\rm beam}^*$ obtaining much better performance. A possible explanation for the performance is that the population-based stable and greedy phase shift of the homogeneous RIS elements carefully locates the scattered optimums in which the global one may exist. The explanation also suits ILS, which obtains the second-best results on most instances. In comparison, the representative sequential beamforming and other baselines underperform. All these demonstrate ALDes's efficiency. 

\begin{table*}[t]
\centering
\caption{Performance (mean$\pm$std) over all runs on test instances of the beamforming problem. $Alg_{\rm beam}^*$'s performance is averaged over 5 algorithms inferred from ALDes's 5 training trials with different seeds. Best results according to pair-wise Wilcoxon sign test ($5\%$ significance level) are with grey background.}
\label{tab_beam}
\begin{tabular}{lccccc}
\toprule
                       & \multicolumn{5}{c}{Number of RIS elements in the test instances}\\
                       & 120 & 160 & 280 & 320 & 400\\
\midrule
Random                 & 0.0442$\pm$7.94E-04 & 0.0425$\pm$6.56E-04 & 0.0402$\pm$8.30E-04 & 0.0390$\pm$6.67E-04 & 0.0375$\pm$1.88E-04 \\
Sequential             & 0.0382$\pm$6.19E-04 & 0.0387$\pm$6.75E-04 & 0.0374$\pm$4.17E-04 & 0.0369$\pm$4.38E-04 & 0.0354$\pm$8.27E-04 \\
ILS                    & 0.0333$\pm$3.74E-04 & 0.0314$\pm$2.49E-04 & 0.0285$\pm$1.19E-04 & 0.0279$\pm$1.82E-04 & 0.0278$\pm$1.15E-04 \\
SA                     & 0.0398$\pm$5.59E-04 & 0.0388$\pm$7.75E-04 & 0.0369$\pm$3.27E-04 & 0.0360$\pm$4.18E-04 & 0.0355$\pm$9.50E-04 \\
GA                     & 0.0369$\pm$3.30E-04 & 0.0356$\pm$1.00E-04 & 0.0337$\pm$4.26E-04 & 0.0333$\pm$1.04E-04 & 0.0322$\pm$6.96E-04 \\
$Alg_{\rm beam}^*$ & \cellcolor{gray}0.0331$\pm$4.03E-0 & \cellcolor{gray}0.0312$\pm$1.08E-07 & \cellcolor{gray}0.0281$\pm$1.23E-07 & \cellcolor{gray}0.0272$\pm$2.39E-08 & \cellcolor{gray}0.0259$\pm$4.41E-08 \\
\bottomrule
\end{tabular}
\end{table*}

\begin{algorithm}[t]
\caption{$Alg_{\rm beam}^*$ for Beamforming}
\label{alg_beam}
\footnotesize
\KwIn{initial solutions $S$}
\While{stopping criterion not met}{		
$S=\texttt{tournament}(S)$ \\			
$S_{new}=\texttt{cross\_n}(S,n=1)$ \\			
$S_{new}=\texttt{reset\_n}(S_{new},n=1)$ \\
$S=\texttt{greedy\_select}(S,S_{new})$ \\		
}
\KwOut{optimal phase shifts from $S$}
\end{algorithm}

\subsection{Application to Power System Restoration} \label{sec_app1}
\subsubsection{Problem Description}
Power system restoration involves scheduling black start (BS) and non-black start (NBS) units to re-energize the power network and loads after a complete blackout, ensuring system safety and stability. The restoration process consists of BS units restart followed by NBS units reboot, receiving cranking power from BS units. Critical loads are then re-energized to maintain system frequency and voltage. Subsequently, remaining loads are re-energized based on priority. Decision variables include restoration sequence and path for BS units, NBS units, and loads, aiming to minimize the total restoration time:
\begin{subequations}
	\label{eq_restore}
	\begin{align}
        &\min_{\mathbf{v}_a,\mathbf{h}_b}\quad\sum_{a=1}^{A}\mathbf{v}_{a}^{T}\mathbf{w}_{a}+\sum_{b=1}^{B}\mathbf{h}_{b}^{T}\mathbf{l}_{b}, \\ 
        &\quad s.t.\quad \sum_{c=1}^{C}v_{ca}=1, a=1,2,\cdots,A, \\ 
        &\qquad\quad\    \sum_{a=1}^{A}h_{ab}=1, b=1,2,\cdots,B, \\
        &\qquad\quad\    0\leq P_{i}^{gen}\leq P_{i}^{gen,max}, \forall i\in\Omega_{Gen}, \\
        &\qquad\quad\    Q_{i}^{gen,min}\leq Q_{i}^{gen}\leq Q_{i}^{gen,max}, \forall i\in\Omega_{Gen}, \\
        &\qquad\quad\    0\leq P_{a,t-1}^{gen}-P_{a,t}^{gen}\leq R_{G}\Delta t, \forall a\in\Omega_{NBS}, \\
        &\qquad\quad\    P_{m,t}=\sum_{mn\in\Omega_{Br}}P_{mn,t}, \\
        &\qquad\quad\    Q_{m,t}=\sum_{mn\in\Omega_{Br}}Q_{mn,t}, \\
        &\qquad\quad\    P_{mn}^{2}+Q_{mn}^{2}\leq S_{mn}^{2}z_{mn}, \forall mn\in\Omega_{Br}, \\
        &\qquad\quad\    U_{m}^{min}\leq U_{m}\leq U_{m}^{max}, \forall m\in\Omega_{Bus}, \\
        &\qquad\quad\    \sum_{t=1}^{T}(1-z_{mc,t})\geq t_{c}^{start}, \forall mc\in\Omega_{Br\_BS},\forall c\in\Omega_{BS}, \\
        &\qquad\quad\    z_{ma,t}\leq z_{a,t}\leq\sum_{ma\in\Omega_{Br\_NBS}}z_{ma,t}, \forall a\in\Omega_{NBS},
	\end{align}
\end{subequations}
where binary vectors $\mathbf{v}_{a}=(v_{1a},\cdots,v_{ca},\cdots,v_{Ca})$, $a=1,2,\cdots,A$, and $\mathbf{h}_{b}=(h_{1b},\cdots,h_{ab},\cdots,h_{Ab})$, $b=1,2,\cdots,B$, are decision variables that determine the connectivity from BS units to NBS units and that from NBS units to loads, respectively; $A$, $B$, and $C$ are the numbers of NBS units, loads, and BS units, respectively; $\mathbf{v_{a}}$ donates which BS unit supplies cranking power to NBS unit $a$; $\mathbf{h_{b}}$ determines which NBS unit supplies electricity to load $b$; $\mathbf{w_{a}}$ and $\mathbf{l_{b}}$ stand for the distances of the shortest time path from BS units to NBS units and that from NBS units to loads, respectively; $P_{i}^{gen}$, $P_{m,t}$, and $P_{mn}$ are the active power of generator $i$, bus $m$, and branch $mn$, respectively; $Q_{i}^{gen}$, $Q_{m,t}$ and $Q_{mn}$ are the reactive power of generator $i$, bus $m$, and branch $mn$, respectively; $P_{i}^{gen,max}$ is the maximum active power of generator $i$; $Q_{i}^{gen,min}$ and $Q_{i}^{gen,max}$ are the minimum and maximum reactive power of generator $i$, respectively; $R_{G}$ is generators' ramp rate; $S_{mn}$ is the apparent power of branch $mn$; $U_{m}$ is the voltage of bus $m$; $U_{m}^{min}$ and $U_{m}^{max}$ are the minimum and maximum voltage of bus $m$, respectively; $t_{c}^{start}$ is the time that BS unit $c$ starts generating power; $z_{mn}$, $z_{mc,t}$, $z_{ma,t}$, and $z_{a,t}$ are the restore status of branch $mn$, that of branch $mc$ that directly connects to BS unit $c$ at time $t$, that of branch $ma$ that directly connects to NBS unit $a$ at time $t$, and that of NBS unit $a$ at time $t$, respectively.

Equation \eqref{eq_restore} represents a binary integer programming problem with highly non-convex constraints, typically NP-hard. Attaining a feasible solution satisfying these constraints using conventional analytical solvers is challenging and computationally intensive as the power system scale increases. Metaheuristics offer a potential solution.

\subsubsection{Applying ALDes} 
We apply ALDes to generate metaheuristic algorithms for the power system restoration problem. Two instances of the IEEE 39-bus system are implemented. They differ in the locations of BS and NBS units. Baselines include the analytical solver CPLEX, ILS, and GA. Table \ref{tab_restoration} reports the results on the test instance, in which $Alg_{\rm restor}^*$ is the algorithm generated by ALDes. All metaheuristic peers handle the constraints by penalizing solutions' fitness with constraint violations. The results measure the restoration time; smaller values refer to shorter restoration time.

$Alg_{\rm restor}^*$ obtains much better result than baselines. It performs population-based local search (Algorithm \ref{alg_restoration}). That is, every current solution is chosen to search from (\texttt{traverse}), resulting in a new solution; the current solution is paired with the new one, and the better between them is selected (\texttt{pairwise\_select}). The population of local search pathways are interconnected through uniform crossover. The key difference of $Alg_{\rm restor}^*$ from metaheuristic baselines is its diversity maintenance by many local search pathways, enabling it to explore feasible and promising solutions within the fragmented solution space. CPLEX lags behind metaheuristic solvers, because it smooths the fragmented space by relaxation, potentially missing the true optimal. All these demonstrate metaheuristics' significance in practical problem-solving and ALDes's ability to customize metaheuristic algorithms.

\begin{table}[t]
\centering
\caption{Performance (mean$\pm$std) on the test instance of the power system restoration problem. $Alg_{\rm restor}^*$'s performance is averaged over 5 algorithms inferred from ALDes's 5 training trials with different seeds. Best results according to pair-wise Wilcoxon sign test ($5\%$ significance level) are with grey background.}
\label{tab_restoration}
\begin{tabular}{p{4cm}p{4cm}}
\toprule
Algorithms                & Restoration Time      \\ 
\midrule
CPLEX                     & 3697.82$\pm$0.00E+00   \\
ILS                       & 3268.51$\pm$1.43E+03   \\
GA                        & 3194.87$\pm$2.38E+02   \\
$Alg_{\rm restor}^*$        & \cellcolor{gray}3016.52$\pm$2.24E+01 \\     
\bottomrule
\end{tabular}
\end{table}

\begin{algorithm}[t]
\caption{$Alg_{\rm restor}^*$ for Power System Restoration} 
\label{alg_restoration}
\footnotesize
\KwIn{initial solutions $S$}
\While{stopping criterion not met}{		
    $S=\texttt{traverse}(S)$ \\			
    $S_{new}=\texttt{cross\_uniform}(S,p=0.2)$ \\			
    $S_{new}=\texttt{reset\_n}(S_{new},n=1)$ \\
    $S=\texttt{pairwise\_select}(S,S_{new})$ \\		
}
\KwOut{best solution from $S$}
\end{algorithm} 

\section{Conclusions}\label{sec_conclusion}
This paper has presented the autoregressive learning-based designer, ALDes, for automated design of metaheuristic algorithms. ALDes formulates metaheuristic algorithm design as a sequence generation task and proposes a sequence representation of metaheuristic algorithms. The new formulation and representation push the boundary of automated algorithm design to a new level: designing algorithms with diverse structures that can fully discover potentials over the metaheuristic family. ALDes leverages an autoregressive generative network with a transformer backbone to handle the task. The network is trained offline to infer high-performance algorithms to subsequently solve many problem instances from the target problem domain. Extensive experiments have revealed that ALDes can customize various types of algorithms (\eg, VLS-, ILS-, GA-style) and even unconventional ones for different problem-solving contexts. The generated algorithms have suited well corresponding target problems and significantly outperformed human-crafted baselines on 24 out of 25 problems.    

This paper has implemented ALDes with a straightforward model architecture, highlighting the efficiency and promise of the autoregressive generation idea. Future research is worth investigating more complex and larger models to develop a \textit{large pre-trained designer}, which could advance the long pursuit of general, open-ended, and automated problem-solving.  

\bibliographystyle{IEEEtran}
\bibliography{References}

\begin{appendices}
\setcounter{table}{0}   
\setcounter{figure}{0}
\renewcommand{\thetable}{A\arabic{table}}
\renewcommand{\thefigure}{A\arabic{figure}}

\section{Experimental Setup: Baselines} \label{appendix_baseline}
Baselines' setup is shown in Table \ref{tab_settings}.

\begin{table}[htbp]
\centering
\scriptsize
\caption{Baseline Settings}
\label{tab_settings}
\begin{tabular}{p{2cm}p{6cm}}
\toprule
Baselines & Settings \\
\midrule
ILS  & Random restart as global perturbation; "no improvement in three consecutive iterations" as acceptance criterion. \\
SA   & Initial temperature enables the initial acceptance rate of $0.8$; cooling schedule of $0.995$.  \\
TS   & Tabu list length$=10\%\times d$; Tabu tenure$=10\%\times d$. \\
\bottomrule
\end{tabular}
\end{table}

\section{Experimental Setup: Design Space} \label{appendix_space}
According to Section \ref{sec_sequence}, the design space contains four types of tokens, \ie, \textit{components}, \textit{hyperparameters}, \textit{pointers}, and \textit{conditions}. The \textit{components} employed in the experiments and applications is shown in Table \ref{tab_space}. They are deconstructed into three categories. The first is Choose components that work on choosing certain current solutions to search from. Among them, \texttt{traverse} makes the input and output identical, as done (implicitly) in most individual- and swarm-based metaheuristics; \texttt{roulette\_wheel}, \texttt{tournament}, and \texttt{nich} output a segmentation or augmentation of the input, as done by evolutionary algorithms' mating selections. The second is Search components that conduct stochastic search based on the chosen solutions. The third is Select components that select promising solutions from the current ones. We use these essential components rather than sophisticated ones, in order to make a fair comparison with baselines and ensure the investigation on the proposed ALDes is not deceived by specific design choices. In practice, complex design choices from prior experimentation and domain expertise can be employed to boost the performance on specific problem-solving. 

Some components have \textit{hyperparameters} ($n$, $p$ in Table \ref{tab_space}). To fit for the one-hot conversion in the sequence embedding module of ALDes (Figure \ref{fig_designer}), we consider 10 choices of $n$, \ie, $\{1\%,5\%,10\%,\cdots\}\times d$ with a step size of $5\%$. $p$ is discretized into also 10 choices $\{1\%,5\%,10\%,\cdots\}$ with a step size of $5\%$. This results in 10 candidates for a hyperparameter token. 

Choices of \textit{pointers} are detailed in Section \ref{sec_sequence}. For \textit{conditions}, we consider 5 count-conditions, \ie, $\{1\%,5\%,10\%,15\%,20\%\}\times$FEs.  This results in 5 candidates for a condition token. For generality over the experiments and applications, event-conditions are not involved. 

\begin{table}[htbp]
\centering
\scriptsize
\caption{Metaheuristic algorithm components in the design space.}
\label{tab_space}
\begin{tabular}{p{3.5cm}p{4.5cm}}
\toprule
Component & Description \\
\midrule
\multicolumn{2}{l}{Choose where to search from:} \\
\texttt{traverse}             & Choose each of the current solution(s) \\
\texttt{roulette\_wheel}      & Choose by roulette wheel\\
\texttt{tournament}           & Choose by tournament \\
\texttt{nich}                 & Choose by adaptive niching \\                           
Search:              &  \\
\texttt{reset\_n}             & Reset $n$ randomly selected entities \\
\texttt{reset\_rand}          & Reset each entity to a random value with probability $p$\\
\texttt{reset\_creep}         & Add a small positive or negative value to each entity with probability $p$, for problems with ordinal attributes \\ 
\texttt{cross\_n}             & $n$-point crossover \\
\texttt{cross\_uniform}       & Uniform crossover with probability $p$\\
\\
\texttt{reinitialize}         & Random restart \\ 
\multicolumn{2}{l}{Select:} \\
\texttt{greedy\_select}               & Select the best solutions \\
\texttt{pairwise\_select}             & Select the better solution from each pair of old and new solutions \\
\texttt{round\_robin\_select}         & Select solutions by round-robin tournament \\
\texttt{simulated\_annealing\_select} & Select according to the Metropolis condition \\
\texttt{tabu}                 & The tabu list \\
\bottomrule
\end{tabular}
\end{table}

\section{Experimental Setup: Problem Embedding} \label{appendix_problem}
32 landscape features are chosen as factors for the problem embedding module of ALDes (Figure \ref{fig_designer}), following \cite{renau2020exploratory}. The factors are summarized in Table \ref{tab_factor}; detailed descriptions please refer to Flacco \cite{kerschke2019comprehensive}. For each problem, we sample $100\times d$ solutions by random walk sampling \cite{malan2013survey} for calculating the factor values. Hamming distance is employed to measure similarity in the factor calculation. We do 5 trials of sampling and factor calculation with different seeds, and average the values, to alleviate uncertainty. We choose these factors because they are computationally cheap and independent of sampling methods \cite{renau2020exploratory}. Note that the sampling does not incur additional computational overhead, as we randomly select solutions from the sampled ones and reuse them as initial solutions of the generated algorithm, when estimating the algorithm's expected performance $R$ (Equation \eqref{eq_train1}). Furthermore, since we use the factors to recognize problems rather than analyze the landscape, the choose of factors matters little. 

\begin{table}[htbp]
\centering
\scriptsize
\caption{Factors for problem embedding.}
\label{tab_factor}
\begin{tabular}{p{4cm}p{4cm}}
\toprule
disp.ratio\_mean\_02                    & ela\_meta.quad\_simple.adj\_r2      \\
disp.ratio\_mean\_05                    & ela\_meta.quad\_simple.cond         \\
disp.ratio\_mean\_10                    & ela\_meta.quad\_w\_interact.adj\_r2 \\
disp.ratio\_mean\_25                    & ela\_meta.costs\_runtime            \\
disp.ratio\_median\_02                  & ic.h\_max                           \\
disp.ratio\_median\_05                  & ic.eps\_s                           \\
disp.ratio\_median\_10                  & ic.eps\_max                         \\
disp.ratio\_median\_25                  & ic.eps\_ratio                       \\
disp.diff\_mean\_02                     & ic.m0                               \\
disp.diff\_mean\_05                     & ic.costs\_runtime                   \\
ela\_meta.lin\_simple.adj\_r2           & nbc.nn\_nb.sd\_ratio                \\
ela\_meta.lin\_simple.intercept         & nbc.nn\_nb.mean\_ratio              \\
ela\_meta.lin\_simple.coef.min          & nbc.nn\_nb.cor                      \\
ela\_meta.lin\_simple.coef.max          & nbc.dist\_ratio.coeff\_var          \\
ela\_meta.lin\_simple.coef.max\_by\_min & nbc.nb\_fitness.cor                 \\
ela\_meta.lin\_w\_interact.adj\_r2      & nbc.costs\_runtime                  \\
\bottomrule
\end{tabular}
\end{table}
\end{appendices}

\iffalse
\begin{IEEEbiography}{Michael Shell}
Biography text here.
\end{IEEEbiography}

% if you will not have a photo at all:
\begin{IEEEbiographynophoto}{John Doe}
Biography text here.
\end{IEEEbiographynophoto}

% insert where needed to balance the two columns on the last page with
% biographies
%\newpage

\begin{IEEEbiographynophoto}{Jane Doe}
Biography text here.
\end{IEEEbiographynophoto}
\fi

\end{document}